# Analytical Comparison of Noise Reduction Filters for Image Restoration Using SNR Estimation

Poorna Banerjee Dasgupta

*M.Tech Computer Science & Engineering, Nirma Institute of Technology*
*Ahmedabad, Gujarat, India*

*Abstract*— Noise removal from images is a part of image restoration in which we try to reconstruct or recover an image that has been degraded by using a priori knowledge of the degradation phenomenon. Noises present in images can be of various types with their characteristic Probability Distribution Functions (PDF). Noise removal techniques depend on the kind of noise present in the image rather than on the image itself. This paper explores the effects of applying noise reduction filters having similar properties on noisy images with emphasis on Signal-to-Noise-Ratio (SNR) value estimation for comparing the results.

## I. INTRODUCTION

Digital images are prone to a variety of types of noise [1],[2]. Noise is the result of errors in the image acquisition process that result in pixel values that do not reflect the true intensities of the real scene. There are several ways in which noise can be introduced into an image, depending on how the image is created. For example:

- If the image is scanned from a photograph made on film, the film grain is a source of noise. Noise can also be the result of damage to the film, or be introduced by the scanner itself.
- If the image is acquired directly in a digital format, the mechanism for gathering the data (such as a CCD detector) can introduce noise.
- Electronic transmission of image data can introduce noise.

### A. Types of noises in Images

Image degradation maybe caused due to various categories of noises such as: Gaussian, Rayleigh, Erlang, Uniform, Exponential, Salt, Pepper, Salt-and-Pepper noises [1]. In subsequent sections of this paper, three particular categories of noises viz. *Salt, Pepper, Salt-and-Pepper* noises have been studied and comparatively analysed through application of various noise reduction filters. Each result has then been qualitatively assessed with the help of SNR estimation to determine which kind of filter is best suited for removal of a particular noise type when there is a choice among the filters to be used.

Salt-and-pepper noise is also known as bipolar impulse noise. Its characteristic Probability Distribution Function (PDF) is shown in Figure 1[1]. Bipolar impulse noise is specified as:

$$p(z) = \begin{cases} P_a & z = a \\ P_b & z = b \\ 0 & \text{otherwise} \end{cases}$$

Here $z$ represents intensity values of pixels in a noisy image. If $b>a$, intensity $b$ will appear as a light dot on the image and $a$ appears as a dark dot. If either $P_a$ or $P_b$ is zero the noise is called unipolar. Frequently, $a$ and $b$ are saturated values, resulting in positive impulses being white and negative impulses being black.

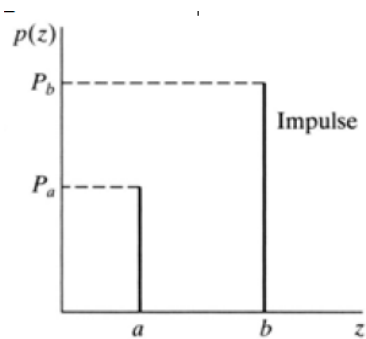

Fig. 1 Impulse bipolar noise

## II. SNR ESTIMATION

There exist many approaches for estimation of the Signal-to-Noise Ratio (SNR) depending on the type of data that is being analysed [3][4][5][6]. However, in the context of digital image processing where all data values are in terms of luminance and are positive values, the most common approach for determining the SNR value is to take the ratio of the mean image pixel intensity values (μ) and the standard deviation of the image pixel values (σ), i.e. SNR = μ/σ. In subsequent sections of this paper, this approach for SNR estimation has been used for qualitative analysis and comparison of the outputs of noise reduction filters – higher SNR values are indicative of better noise removal.

III. COMPARISON OF NOISE REDUCTION FILTERS

In this section, a comprehensive comparative study of noise reduction filters with input test images has been carried out. The results and findings of the study have been summarized in Tables 1 to 3. The original, noise-free input test image is shown in Figure 2 [1]. For the original, noise-free image, the following statistics were obtained: $\mu=150.8$, $\sigma=15.4$, SNR $=9.8$.

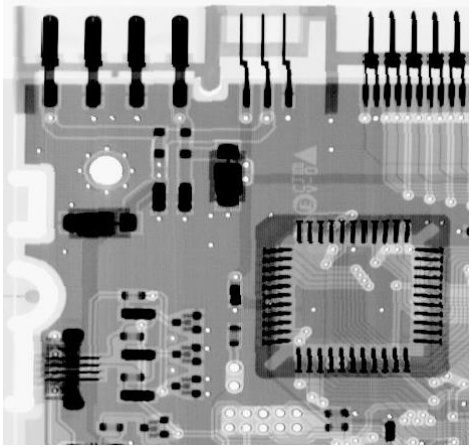

Fig. 2 Original noise-free image

*A. Removal of Salt Noise*

Filters used for noise reduction:
- Min Filter
- Contra-harmonic mean filter (CHM)

The resultant images are shown below. Figure 3(a) shows the input test image with salt noise [1]. Figure 3(d) shows the output after applying a 3x3 Min filter and Figure 3(e) shows the result of subtracting this output from the input test image. Figure 3(b) shows the output after applying Contra-harmonic mean filter with Q-parameter = -1 and Figure 3(c) shows the result of subtracting this output from the input test image. These subtracted images show an estimate of how close the output is with the input image and also the amount of noise removed from the image. SNR values were calculated for each output and the following results were obtained as shown in Table 1.

TABLE I
SNR FOR SALT-NOISE REDUCTION FILTERS

| SNR of input noisy image | SNR of Min Filter's output | SNR of CHM Filter's output |
|---|---|---|
| $\mu=161.2$ $\sigma=16.8$ SNR = 9.6 | $\mu=136.5$ $\sigma=16.1$ SNR=8.5 | $\mu=138.6$ $\sigma=18.2$ SNR=7.6 |

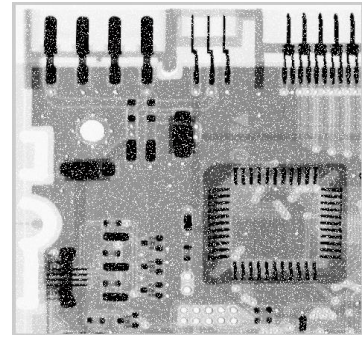

Fig 3(a) Input image with salt noise

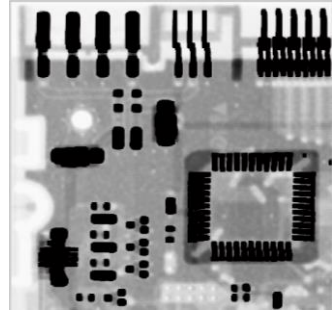 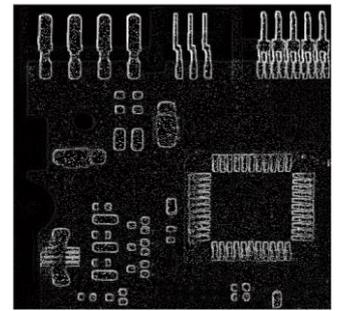

Fig 3(b) Output of CHM Filter    Fig 3(c) Input minus Output of CHM Filter

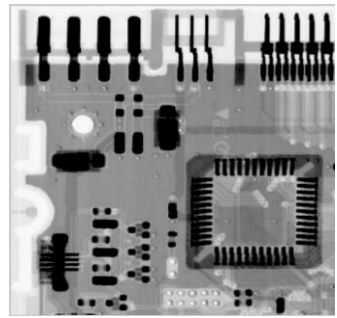 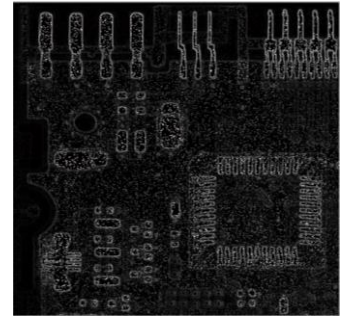

Fig 3(d) Output of Min Filter    Fig 3(e) Input minus Output of Min Filter

Due to presence of salt noise in the image, the mean value comes to be quite high. However, after applying the filters, it has been found that the Min filter yields a better result and a closer SNR value to that of the original noise-free image. Also it was noticed that applying Contra-harmonic mean filters leads to undesirable thickening of dark areas in the image. This is especially noticeable for the pins in the figure of the circuit diagram.

*B. Removal of Pepper Noise*

Filters used for noise reduction:
- Max Filter
- Contra-harmonic mean filter (CHM)

The resultant images are shown below. Figure 4(a) shows the input test image with pepper noise [1]. Figure 4(b) shows the output after applying a 3x3 Max filter and Figure 4(c) shows the result of subtracting this output from the input test image. Figure 4(d) shows the output after applying Contra-harmonic mean filter with Q-parameter = +1 and Figure 4(e) shows the result of subtracting this output from the input test image. SNR values were calculated for each output and the following results were obtained as shown in Table 2.

TABLE II
SNR FOR PEPPER-NOISE REDUCTION FILTERS

| SNR of input noisy image | SNR of Max Filter's output | SNR of CHM Filter's output |
|---|---|---|
| μ=135.7 σ=8.2 SNR =16.4 | μ=165.0 σ=16.1 SNR=10.2 | μ=161.4 σ=13.4 SNR=12.1 |

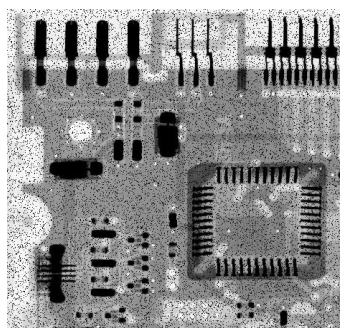

Fig 4(a). Input image with pepper noise

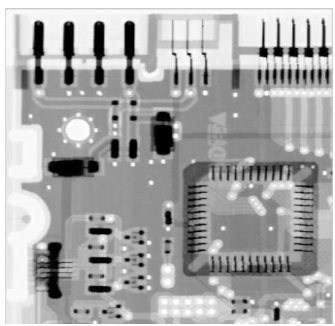 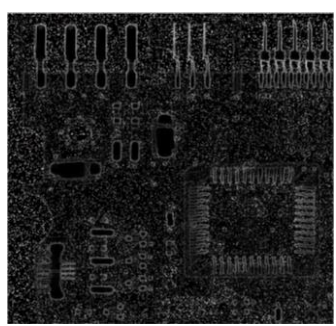

Fig 4(b) Output of Max Filter    Fig 4(c). Output of Max Filter minus Input

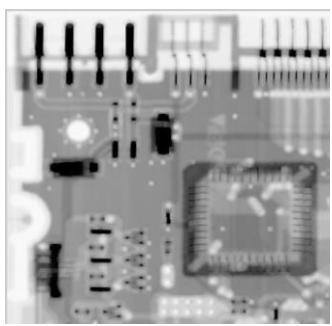 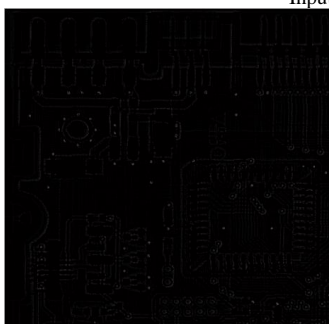

Fig 4(d). Output of CHM Filter    Fig 4(e). Input minus Output of CHM Filter

Due to presence of pepper noise in the image, the mean value comes to be lower than that of the original noise-free image. However, after applying the filters, it has been found that the Max filter yields a better result and a closer SNR value to that of the original noise-free image. Also it was noticed that applying Contra-harmonic mean filters leads to a higher SNR value but also produces an undesirable "washed-out" effect.

### C. Removal of Salt-and-Pepper Noise

Filters used for noise reduction:
- Static Median Filter (SMF)
- Adaptive Median Filter (AMF)

The resultant images are shown below. Figure 5(a) shows the input noisy test image [1]. Figure 5(b) shows the output after applying a 3x3 Static Median filter and Figure 5(c) shows the result of subtracting this output from the input test image. Figure 5(d) shows the output after applying Adaptive Median filter with maximum allowable filter size of 5x5 and Figure 5(e) shows the result of subtracting this output from the input test image. SNR values were calculated for each output and the following results were obtained as shown in Table 3.

TABLE III
SNR FOR SALT-PEPPER NOISE REDUCTION FILTERS

| SNR of input noisy image | SNR of Static Median Filter's output | SNR of Adaptive Median filter's output |
|---|---|---|
| μ=144.1 σ=7.7 SNR=18.7 | μ=148.2 σ=15.2 SNR=9.7 | μ=146.9 σ=13.8 SNR=10.7 |

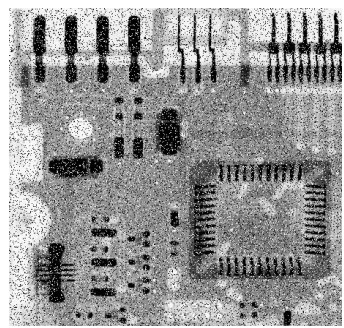

Fig 5(a) Input image with salt and pepper noise

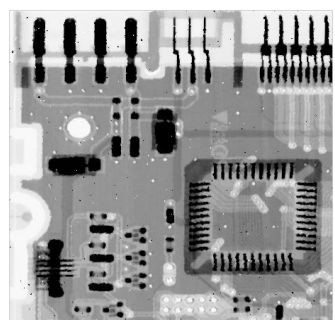 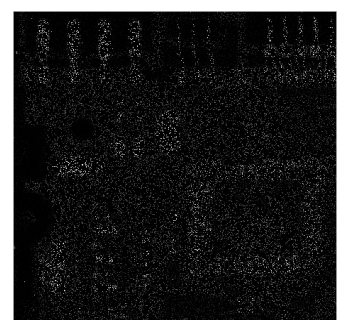

Fig 5(b) Output of SMF    Fig 5(c) Input minus Output of SMF

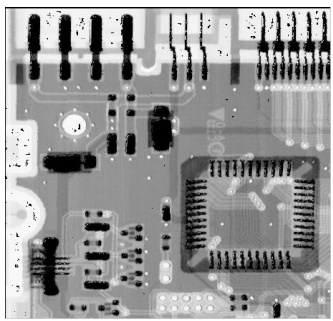
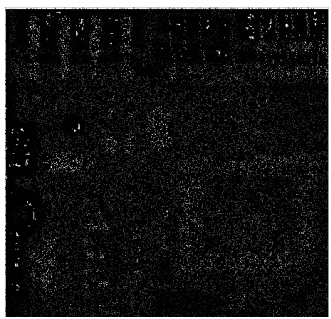

Fig 5(d) Output of AMF     Fig 5(e) Input minus Output of AMF

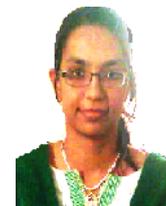

AUTHOR'S PROFILE:
**Poorna Banerjee Dasgupta** has received her B.Tech & M.Tech Degrees in Computer Science and Engineering from Nirma Institute of Technology, Ahmedabad, India. She did her M.Tech dissertation at Space Applications Center, ISRO, Ahmedabad, India and has also worked as Assistant Professor in Computer Engineering dept. at Gandhinagar Institute of Technology, Gandhinagar, India from 2013-2014. Her research interests include image processing, high performance computing, parallel processing and wireless sensor networks.

Due to presence of both pepper and salt noise in the image, the mean value comes to be quite close to that of the original noise-free image. However, after applying the filters, it has been found that the Static Median filter yields a better result and a closer SNR value to that of the original noise-free image. Also it was noticed that applying Adaptive Median filters leads to a higher SNR value but also produces undesirable black boundaries if zero-padding is used for border pixels. Also it is more time consuming than applying Static Median filters. However using Adaptive Median Filters help preserve edges better which are a part of the original image.

IV. CONCLUSIONS & FUTURE SCOPE OF WORK

Noises present in images can be of various types with their characteristic probability distribution functions. Noise removal techniques depend on the kind of noise present in the image rather than on the image itself. This paper explored the effects of applying noise filters having similar effects on noisy images with emphasis on SNR value estimation for comparing the results. Three categories of noises were analysed viz. Salt noise, Pepper noise and Salt-&-Pepper noise. For each type of noisy image, different filters were applied for noise removal and the filter outputs were then qualitatively assessed using SNR values of each output.

As further extensions to the research work carried out in this paper, more filters can be analysed for other categories of noises and other quality parameters such as edge restoration in images can be used to assess the filter outputs. Also the analysis can be further extended to color images as well.

REFERENCES

[1] Rafael C. Gonzalez ,Richard E. Woods. *Digital Image Processing*, 3rd Edition, Prentice Hall Publications, 2000.
[2] Peter Kellman, Elliot R. McVeigh. *Image reconstruction in SNR units: A general method for SNR measurement*. Wiley Publications, 2005.
[3] John C. Russ. *The image processing handbook*. 5th Edition CRC Press,2007.
[4] Suk Hwan Lim ; Maurer, R. ; Kisilev, P. "Denoising scheme for realistic digital photos from unknown sources". IEEE International Conference on Acoustics, Speech and Signal Processing, 2009.
[5] D. J. Schroeder. *Astronomical Optics* ,2nd Edition, Academic Press, 1999.
[6] Tania Stathaki. *Image fusion: algorithms and applications*. Academic Press, 2008.